\pdfoutput=1

\documentclass[11pt]{article}

\usepackage[]{EMNLP2023}

\usepackage[T1]{fontenc}

\usepackage[utf8]{inputenc}

\usepackage{microtype}

\usepackage[ruled,noend]{algorithm2e}
\usepackage{amsmath,amsfonts,amssymb,amsthm}
\usepackage{times}
\usepackage{latexsym}
\usepackage{amsmath,amsfonts,amssymb,amsthm,mathtools}
\usepackage{comment}
\usepackage{graphicx}
\usepackage{xcolor}
\usepackage{color,soul}
\usepackage{booktabs}
\usepackage{mathtools}
\usepackage{linguex}
\usepackage{subfigure}
\usepackage{array}
\usepackage{float}
\usepackage{bm}
\usepackage{arydshln}
\usepackage{tikz}
\usepackage{multirow}
\usepackage{hyperref}
\usepackage{algpseudocode}
\usepackage{pmboxdraw}

\usepackage{inconsolata}

\title{LM Transparency Tool:\\ Interactive Tool for Analyzing Transformer Language Models}

\usepackage{comment}
\usepackage{todonotes}

\definecolor{metacolor}{HTML}{2e72d9}

\author{Igor Tufanov\textcolor{metacolor}{$^{\infty}$} \quad Karen Hambardzumyan\textcolor{metacolor}{$^{\infty}$}$^{\square}$ \quad  Javier Ferrando$^{\diamondsuit}$\thanks{\, \,Work done during an internship at Meta.}  \quad Elena Voita\textcolor{metacolor}{$^{\infty}$}  \\
\textcolor{metacolor}{$^{\infty}$}AI at Meta (FAIR) ~ $^{\square}$University College London ~ $^{\diamondsuit}$Universitat Politècnica de Catalunya\\ 
 \texttt{\{igortufanov,mahnerak,lenavoita\}@meta.com}\\
 \texttt{javier.ferrando.monsonis@upc.edu}}
\begin{document}
\maketitle
\begin{abstract}

We present the LM Transparency Tool (\texttt{LM-TT}), an open-source interactive toolkit for analyzing the internal workings of Transformer-based language models. Differently from previously existing tools that focus on isolated parts of the decision-making process, our framework is designed to make the entire prediction process transparent, and allows tracing back model behavior from the top-layer representation to very fine-grained parts of the model. Specifically, it (i)~shows the important part of the whole input-to-output information flow, (ii)~allows attributing any changes done by a model block to individual attention heads and feed-forward neurons, (iii)~allows interpreting the functions of those heads or neurons. A crucial part of this pipeline is showing the importance of specific model components at each step. As a result, we are able to look at the roles of model components only in cases where they are important for a prediction. Since knowing which components should be inspected is key for analyzing large models where the number of these components is extremely high, we believe our tool will greatly support the interpretability community both in research settings and in practical applications. We make \texttt{LM-TT} codebase available at \url{https://github.com/facebookresearch/llm-transparency-tool}

.

\end{abstract}

\section{Introduction}

Recent advances in natural language processing led to remarkable capabilities of the Transformer language models, especially with scale~\citep{NEURIPS2020_1457c0d6,kaplan2020scaling,opt_lm,wei2022emergent,ouyang2022training,openai2023gpt4,anil2023palm,touvron2023llama,touvron2023llama2}. This, along with the wide adoption of such models in high-stakes settings, makes understanding the internal workings of these models vital from the safety, reliability and trustworthiness perspectives.

\begin{figure*}[!t]
	\begin{centering}
	\includegraphics[width=1.0\textwidth]{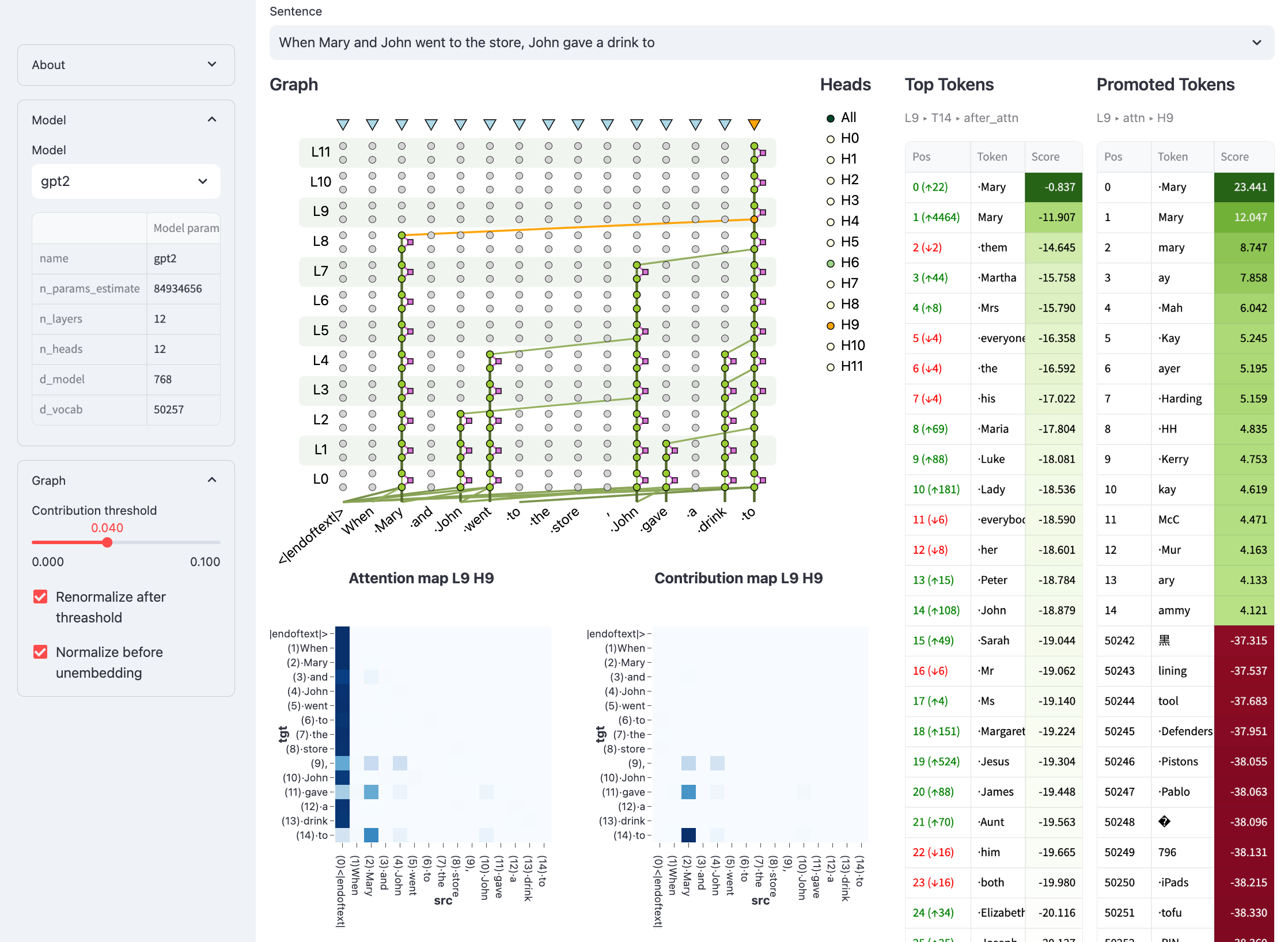}
	\caption{The LM Transparency Tool UI showing information flow graph for the selected prediction, importances of attention heads at the selected layer, attention and contribution maps, logit lens for the selected representation, and top tokens promoted/suppressed by the selected attention head.}
	\label{fig:main}
	\end{centering}
\end{figure*}

Existing tools for analyzing sequence models' predictions enable users to compute input tokens attribution scores, read token promotions performed by different model components, or analyze textual patterns responsible for the activation of model's neurons~\citep{geva-etal-2022-lm,katz-belinkov-2023-visit,alammar-2021-ecco,tenney-etal-2020-language,sarti-etal-2023-inseq,kokhlikyan2020captum,miglani-etal-2023-using}. However, these focus only on specific parts of the decision-making process and none of them is designed to make the entire prediction process transparent. In contrast, we introduce LM Transparency Tool, a framework that allows tracing back model behavior to very fine-grained model parts.

One of the key advantages of our pipeline is the ability to look only at those model components that were relevant for a selected prediction. Indeed, e.g. syntactic attention heads~\cite{voita-etal-2019-analyzing}, induction heads~\cite{elhage2021mathematical,olsson2022context}, knowledge neurons~\cite{dai-etal-2022-knowledge}, etc. perform their function only in specific cases and are ``dormant'' otherwise~-- therefore, looking at them makes sense only for those certain examples. To make this possible, our tool first
shows the information flow routes introduced by~\citet{ferrando_voita2024routes}: this is a subset of intermediate token representations and model components that together form the most important part of the entire input-to-output processing. Then, the tool further
allows (i)~attributing any changes done by those important model blocks to individual attention heads and feed-forward neurons, as well as (ii)~interpreting the functions of those heads and neurons. Importantly, \texttt{LM-TT} is highly efficient: due to relying on~\citet{ferrando_voita2024routes}, it is 100 times faster than typical patching-based alternatives~\cite{conmy2023automated}.

Overall, the LM Transparency Tool:
\begin{itemize}
    \item visualizes the ``important'' part of the prediction process along with importances of model components at varying levels of granularity;
    \item allows interpreting representations and updates coming from model components;
    \item enables analyzing large models where it is crucial to know what to inspect;
    \item allows interactive exploration via a UI;
    \item is highly efficient.
\end{itemize}

\section{User Interface and Functionality}

Inside Transformer language models, each representation evolves from the current input token embedding\footnote{Sometimes, along with positional encoding.} to the final representation used to predict the next token. This evolution happens through additive updates coming from attention and feed-forward blocks. The resulting stack of same-token representations is usually referred to as ``residual stream''~\citep{elhage2021mathematical}, and the overall computation inside the model can be viewed as a sequence of residual streams connected through layer blocks. Formally, we can see it as a graph where nodes correspond to token representations and edges correspond to operations inside the model (attention heads, feed-forward layers, etc.).
Our tool visualizes the ``important'' part of this graph, importances of model components at varying levels of granularity (individual heads and neurons), as well as an interpretation of representations and updates coming from model components.

\subsection{Important Information Flow Subgraph}

As we mentioned, we can see computations inside the Transformer as a graph with token representations as nodes and operations inside the model as edges.
While during model computation all the edges (i.e., model components) are present, 
computations important
for each prediction are likely to form only a small portion of the original graph~(\citet{voita-etal-2019-analyzing,wang2023interpretability,hanna2023does}, among others). Recent work by \citet{ferrando_voita2024routes}
extracts this important subgraph in a top-down manner by tracing information back through the network
and, at each step, leaving only edges that were important~(Figure~\ref{fig:main}). 
To understand which edges are important, they rely on an attribution method~\cite{ferrando-etal-2022-measuring}. \citet{ferrando_voita2024routes} explain the benefits of this method including, among other things, why it is more versatile, informative and around 100 times more efficient compared to commonly used patching-based approaches typical for the existing mechanistic interpretability workflows~\cite{wang2023interpretability,hanna2023does,conmy2023automated, stolfo2023understanding,docstring}.

\paragraph{In the tool.} In the tool, we show only the important attention edges and feed-forward blocks (purple squares in Figure~\ref{fig:main}). Clicking at the top triangles gives the important information flow routes for each token position. Under the ``Graph'' menu, one can vary the importance threshold to get more or less dense graphs.

\subsection{Fine-Grained Importances}

While the information flow graph already relies on the importances of attention or feed-forward blocks for the current residual stream, the tool goes further and shows the importances of (i)~individual attention heads, and (ii)~individual FFN neurons. 

\subsubsection{Individual Attention Heads}

\paragraph{Importance.} After clicking on an attention edge (green lines in Figure~\ref{fig:main}), the tool shows which specific attention head is mostly responsible for this connection, as well as highlights the importances of other heads for this specific step. 

\paragraph{Weights and contributions.} Whenever a head is selected, the tool shows
\begin{itemize}\itemsep0em
    \item attention map,
    \item contribution map.
\end{itemize}
While the attention map can give an idea of the attention head's function~\cite{voita-etal-2019-analyzing,clark-etal-2019-bert,correia-etal-2019-adaptively}, attention weights might not reflect influences properly~(\citet{bastings-filippova-2020-elephant,kobayashi-etal-2020-attention}, among others). Therefore, we also show \textit{contribution map} reflecting the influence of a head-token pair in the overall attention block~\cite{ferrando_voita2024routes}. Note that while attention weights always sum to~1, contributions sum to the overall importance of this attention head at each step. 
As a result, contribution maps can be more sparse, as shown in Figure~\ref{fig:main}.

\subsubsection{Individual FFN Neurons}

When clicking on feed-forward blocks (purple squares), the tool shows the top neurons that contributed at this step. Note that this is different from previous work that either considered
top activated neurons~(\citet{geva-etal-2022-lm,alammar-2021-ecco}, among others) or did not consider neurons at all~(\citet{tenney-etal-2020-language,sarti-etal-2023-inseq}, among others). In contrast, our tool shows top \textit{contributing} neurons and makes it possible to look at the functions of neurons only when they are important, i.e. when they perform their function.

\subsection{Vocabulary Projections}

One of the popular ways to
interpret vector representations is to project them onto the model’s vocabulary space. Our tool does this for (i)~representations in the residual stream and (ii)~the updates coming from specific model components.

\subsubsection{Interpreting Representations}

While to get a prediction, we project the final-layer representation onto the output vocabulary, for interpretation, we can project representations at any point inside the residual stream~-- this is called \textit{logit lens}~\cite{logit_lens}. The resulting sequence of distributions (or top-token predictions) illustrates the decision-making process over the course of the Transformer inference. This is used rather prominently to trace the bottom-up changes in the residual stream~(\citet{alammar-2021-ecco,geva-etal-2021-transformer,geva_2022_transformer, merullo2023mechanism,belrose2023eliciting,din2023jump}, among others).

\paragraph{Tool: click on the circles.} In our tool, circles correspond to residual stream representations after applying each model block, either attention or feed-forward; overall, we have two representations per layer. By clicking at each circle, under ``Top tokens'' the tool shows the projection of this residual state onto output vocabulary.\footnote{Under ``Graph'', one also specifies whether to apply the final layer normalization before projecting onto vocabulary or not.}

\subsubsection{Interpreting Model Components}


We can also project onto vocabulary an update coming from a model component: this shows how this component changes the residual stream and, therefore, gives an interpretation of its behavior.
In this way, we can get concepts \textit{promoted} by this component by looking at top positive projections~(\citet{geva_2022_transformer,dar-etal-2023-analyzing}, inter alia)
or \textit{suppressed concepts} by looking at bottom negative projections~\cite{voita2023neurons}.

\paragraph{Tool: click on the circles and go further.} When you click on a representation from the residual stream, in addition to this representation's logit lens, the tool will also show top promoted and suppressed concepts for the last applied block (either attention or feed-forward). By clicking further, you can also select an individual attention head or feed-forward neuron and get an interpretation at a finer-grained level.

\subsection{Additional Controls}
\label{sect:additional_controls}

For the functionality above, the sidebar to the left has additional controls:
\begin{itemize}
    \item \textbf{Model}: 
    \begin{itemize}
        \item [$\circ$]\texttt{GPT-2}~\cite{radford_language_2019},
        \item[$\circ$] \texttt{OPT}~\cite{zhang2022opt},
        \item[$\circ$] \texttt{Llama-2}~\cite{touvron2023llama2},
        \item[$\diamond$] add your own model (Section~\ref{sect:adding_model});
    \end{itemize}
    \item \textbf{Device}: \texttt{GPU} or \texttt{CPU};
    \item \textbf{Data}: adding custom data or choosing an existing example;
    \item \textbf{Graph}: tuning parameters of the information flow graph, e.g. contribution threshold etc.~\cite{ferrando_voita2024routes}.
\end{itemize}

\subsection{Intended Use Cases}

The tool can help generating or validating hypotheses about model functioning more quickly.
The list of potential use cases contains, but is not limited to the following:
\begin{itemize}
    \item finding model components amplifying biases;
    \item checking whether the model is reasoning via different routes for desired/undesired behavior (e.g., in safety settings);
    \item validating whether e.g. mathematical tasks are solved via computation rather than memorization;
    \item inspecting model behavior for factuality, when hallucinating, etc.
\end{itemize}

\section{System Design and Components}
\label{sect:system_design}

Our application is a web-based toolkit, offering easy and interactive access that is cross-platform compatible. This approach allows users to utilize and share the tool remotely, emphasizing convenience and flexibility.

\subsection{Frontend}
The frontend is developed using \texttt{Streamlit}~\cite{streamlit}, with an additional custom component specifically created for visualization of the Transformer model in the form of a graph. This enhancement was necessary as such complex visualizations are not natively supported by \texttt{Streamlit}'s built-in features. The custom component is built using \texttt{D3.js}~\cite{2011-d3} and integrated with \texttt{React} for managing dynamic content and user interactions.

\subsection{Backend}
Our backend is a single-dispatch, stateless \texttt{Streamlit} program. It includes a caching mechanism to optimize performance for repeated queries. The modeling and tokenization are powered by Hugging Face \texttt{transformers}~\citep{wolf-etal-2020-transformers}\footnote{\url{https://github.com/huggingface/transformers}} library. For capturing model activations and intermediate computations, we use \texttt{TransformerLens}~\citep{nanda2022transformerlens}\footnote{\url{https://github.com/neelnanda-io/TransformerLens}} library as it has hooked wrappers defined for a variety of models.

\subsection{Configuration and Deployment}
Configuration is handled via a \texttt{JSON} file, allowing for customization of parameters such as dataset file access, maximum user string length, the list of available models, a default model and a dataset. An example configuration is shown in Figure~\ref{example_config}.

\begin{figure}[!h]
\input{pict/sample_config.tex}
\caption{An example configuration.}
\label{example_config}
\end{figure}
\noindent In this configuration, model names can be either Hugging Face model identifiers or local paths. Other settings, such as threshold adjustments and computation precision, are directly configurable within the application's user interface, enabling quick switching.

Overall, launching the tool is as easy as:

{
\footnotesize
\begin{verbatim}
streamlit run app.py -- path/to/config.json
\end{verbatim}
}

\subsection{Computations}

For a selected sentence and model, the tool makes the forward pass and uses the following tensors:
\begin{itemize}
    \item intermediate representations: residual stream states before and after each block;
    \item each block's output: the value added to the residual stream by FFN or attention;
    \item attention block internal states: attention weights, per-head block output, token-specific terms in each head's output;
    \item FFN block internal states: neuron activations before and after the activation function.
\end{itemize}
Using this, the tool computes the importances of all the elements (blocks, heads, neurons) and extracts the information flow graph~\cite{ferrando_voita2024routes}. 
Vocabulary projections and importances within a layer are done on-the-fly when a user clicks on an element. 

\paragraph{Supported model sizes.} We tested the tool with models up to \texttt{30b} of parameters. Since for simplicity and ease of debugging we focus on single-node setup, larger models requiring distributed mode might not work in the current version.

\paragraph{Efficiency.} The tool supports automatic mixed precision (\texttt{float16} and \texttt{bfloat16}). This helps to store model parameters and tensors efficiently, thus saving memory in order to accommodate larger models without sacrificing performance. Models are loaded on demand and cached for efficiency. 

\subsection{Outside of the UI}

Outside of the UI, one can access the underlying functionality of the tool programmatically with Python function calls. For example, getting information flow routes requires the following call:

{
\footnotesize
\begin{verbatim}
import llm_transparency_tool as lmtt
from lmtt.models.tlens_model import (
    TransformerLensTransparentLlm,
)

model = TransformerLensTransparentLlm(name)

model.run([sentence])

graph = lmtt.routes.graph.build_full_graph(
    model,
    threshold=threshold,
)
\end{verbatim}
}

\subsection{Adding Your Own Models}
\label{sect:adding_model}

By default, upon installation, the tool supports only the models listed in Section~\ref{sect:additional_controls}. Steps needed for adding a new model depend on whether the model is supported by \texttt{TransformerLens}.

\paragraph{Supported by TransformerLens.} Adding a model supported by \texttt{TransformerLens} model is very simple.
\begin{itemize}
    \item \textbf{Hugging Face weights:} Add model name (as stated in Hugging Face \texttt{transformers}) to the app's configuration \texttt{JSON} file. 

    \item \textbf{Custom weights:} In the \texttt{JSON} configuration file, along with the name of the model, provide the path to the model file.
\end{itemize}

\paragraph{Not supported by TransformerLens.}
In this case, you need to let the tool know how to create proper hooks for the model. 
Our tool is using \texttt{TransformerLens} through an intermediate interface (\texttt{TransparentLlm} class) and you have to implement this interface for your model.

\section{Related Work}

Existing tools for analyzing sequence models' predictions include \texttt{LM-Debugger}~\citep{geva-etal-2022-lm}, \texttt{VISIT}~\citep{katz-belinkov-2023-visit}, \texttt{Ecco}~\citep{alammar-2021-ecco}, \texttt{LIT}~\citep{tenney-etal-2020-language}, \texttt{Inseq}~\citep{sarti-etal-2023-inseq}, and \texttt{Captum}~\citep{kokhlikyan2020captum,miglani-etal-2023-using}. These tools enable users to compute input tokens attribution scores, read token promotions performed by different model components via logit lens, or analyze textual patterns responsible for the activation of the model's neurons. However, these are not able to extract a relevant part of model computations and indicate component importances. To identify parts of the model relevant for some task, a recent trend in mechanistic interpretability is to rely on causal interventions on the computational graph of the model, aka ``activation patching''~\cite{causal_mediation_bias,geiger-etal-2020-neural,geiger2021causal,wang2023interpretability,hanna2023does,conmy2023automated, stolfo2023understanding, docstring}. 
Usually, this process involves the following steps: 1)~selecting a dataset and metric, 2)~manually creating contrastive examples, 3)~searching for important edges in the graph via activation patching. The latter requires running a forward pass per each patched element and uses many patches to explain a single prediction. Although recent approaches aim to automate some parts of this workflow~\citep{conmy2023automated}, the entire process requires a large human effort and involves significant computational costs: this imposes constraints on the tool development and limits its applicability. Differently, 
\texttt{LM-TT} relies on a recent method  by~\citet{ferrando_voita2024routes} which refuses from the patching constraints by relying on attribution to define the importances. Furthermore, \texttt{LM-TT} incorporates additional functionalities such as showing fine-grained component importances, logit lens analysis at different levels of granularity, and attention visualization not only via attention weights but also via contributions. This enables users to gain a more comprehensive understanding of the functions executed by each component.

\section{Conclusions}

We release the LM Transparency Tool, an open-source toolkit for analyzing Transformer-based language models that allows tracing back model behavior to specific parts of the model. Specifically, it (i)~shows the important part of the whole input-to-output information flow, (ii)~allows attributing any changes done by a model block to individual attention heads and feed-forward neurons, (iii)~allows interpreting the functions of those heads or neurons. Notably, due to the nature of the underlying method, our tool reduces the number of components to be analyzed by highlighting model components that were relevant to the prediction. This greatly simplifies the study of large language models, with potentially thousands of attention heads and hundreds of thousands of neurons to look at. Moreover, the UI accelerates the inspection process, unlike other frameworks that lack this feature. This assists researchers and practitioners in efficiently generating hypotheses regarding the behavior of the model.

\section{Acknowledgments}
We would like to thank Christoforos Nalmpantis, Nicola Cancedda, Yihong Chen, Andrey Gromov and Mostafa Elhoushi for the insightful discussions.

\bibliography{custom}

\begin{thebibliography}{48}
\expandafter\ifx\csname natexlab\endcsname\relax\def\natexlab#1{#1}\fi

\bibitem[{Alammar(2021)}]{alammar-2021-ecco}
J~Alammar. 2021.
\newblock \href {https://doi.org/10.18653/v1/2021.acl-demo.30} {Ecco: An open source library for the explainability of transformer language models}.
\newblock In \emph{Proceedings of the 59th Annual Meeting of the Association for Computational Linguistics and the 11th International Joint Conference on Natural Language Processing: System Demonstrations}, pages 249--257, Online. Association for Computational Linguistics.

\bibitem[{Anil et~al.(2023)Anil, Dai, Firat, Johnson, Lepikhin, Passos, Shakeri, Taropa, Bailey, Chen, Chu, Clark, Shafey, Huang, Meier-Hellstern, Mishra, Moreira, Omernick, Robinson, Ruder, Tay, Xiao, Xu, Zhang, Abrego, Ahn, Austin, Barham, Botha, Bradbury, Brahma, Brooks, Catasta, Cheng, Cherry, Choquette-Choo, Chowdhery, Crepy, Dave, Dehghani, Dev, Devlin, Díaz, Du, Dyer, Feinberg, Feng, Fienber, Freitag, Garcia, Gehrmann, Gonzalez, Gur-Ari, Hand, Hashemi, Hou, Howland, Hu, Hui, Hurwitz, Isard, Ittycheriah, Jagielski, Jia, Kenealy, Krikun, Kudugunta, Lan, Lee, Lee, Li, Li, Li, Li, Li, Lim, Lin, Liu, Liu, Maggioni, Mahendru, Maynez, Misra, Moussalem, Nado, Nham, Ni, Nystrom, Parrish, Pellat, Polacek, Polozov, Pope, Qiao, Reif, Richter, Riley, Ros, Roy, Saeta, Samuel, Shelby, Slone, Smilkov, So, Sohn, Tokumine, Valter, Vasudevan, Vodrahalli, Wang, Wang, Wang, Wang, Wieting, Wu, Xu, Xu, Xue, Yin, Yu, Zhang, Zheng, Zheng, Zhou, Zhou, Petrov, and Wu}]{anil2023palm}
Rohan Anil, Andrew~M. Dai, Orhan Firat, Melvin Johnson, Dmitry Lepikhin, Alexandre Passos, Siamak Shakeri, Emanuel Taropa, Paige Bailey, Zhifeng Chen, Eric Chu, Jonathan~H. Clark, Laurent~El Shafey, Yanping Huang, Kathy Meier-Hellstern, Gaurav Mishra, Erica Moreira, Mark Omernick, Kevin Robinson, Sebastian Ruder, Yi~Tay, Kefan Xiao, Yuanzhong Xu, Yujing Zhang, Gustavo~Hernandez Abrego, Junwhan Ahn, Jacob Austin, Paul Barham, Jan Botha, James Bradbury, Siddhartha Brahma, Kevin Brooks, Michele Catasta, Yong Cheng, Colin Cherry, Christopher~A. Choquette-Choo, Aakanksha Chowdhery, Clément Crepy, Shachi Dave, Mostafa Dehghani, Sunipa Dev, Jacob Devlin, Mark Díaz, Nan Du, Ethan Dyer, Vlad Feinberg, Fangxiaoyu Feng, Vlad Fienber, Markus Freitag, Xavier Garcia, Sebastian Gehrmann, Lucas Gonzalez, Guy Gur-Ari, Steven Hand, Hadi Hashemi, Le~Hou, Joshua Howland, Andrea Hu, Jeffrey Hui, Jeremy Hurwitz, Michael Isard, Abe Ittycheriah, Matthew Jagielski, Wenhao Jia, Kathleen Kenealy, Maxim Krikun, Sneha Kudugunta, Chang
  Lan, Katherine Lee, Benjamin Lee, Eric Li, Music Li, Wei Li, YaGuang Li, Jian Li, Hyeontaek Lim, Hanzhao Lin, Zhongtao Liu, Frederick Liu, Marcello Maggioni, Aroma Mahendru, Joshua Maynez, Vedant Misra, Maysam Moussalem, Zachary Nado, John Nham, Eric Ni, Andrew Nystrom, Alicia Parrish, Marie Pellat, Martin Polacek, Alex Polozov, Reiner Pope, Siyuan Qiao, Emily Reif, Bryan Richter, Parker Riley, Alex~Castro Ros, Aurko Roy, Brennan Saeta, Rajkumar Samuel, Renee Shelby, Ambrose Slone, Daniel Smilkov, David~R. So, Daniel Sohn, Simon Tokumine, Dasha Valter, Vijay Vasudevan, Kiran Vodrahalli, Xuezhi Wang, Pidong Wang, Zirui Wang, Tao Wang, John Wieting, Yuhuai Wu, Kelvin Xu, Yunhan Xu, Linting Xue, Pengcheng Yin, Jiahui Yu, Qiao Zhang, Steven Zheng, Ce~Zheng, Weikang Zhou, Denny Zhou, Slav Petrov, and Yonghui Wu. 2023.
\newblock \href {http://arxiv.org/abs/2305.10403} {Palm 2 technical report}.

\bibitem[{Bastings and Filippova(2020)}]{bastings-filippova-2020-elephant}
Jasmijn Bastings and Katja Filippova. 2020.
\newblock \href {https://doi.org/10.18653/v1/2020.blackboxnlp-1.14} {The elephant in the interpretability room: Why use attention as explanation when we have saliency methods?}
\newblock In \emph{Proceedings of the Third BlackboxNLP Workshop on Analyzing and Interpreting Neural Networks for NLP}, pages 149--155, Online. Association for Computational Linguistics.

\bibitem[{Belrose et~al.(2023)Belrose, Furman, Smith, Halawi, Ostrovsky, McKinney, Biderman, and Steinhardt}]{belrose2023eliciting}
Nora Belrose, Zach Furman, Logan Smith, Danny Halawi, Igor Ostrovsky, Lev McKinney, Stella Biderman, and Jacob Steinhardt. 2023.
\newblock \href {http://arxiv.org/abs/2303.08112} {Eliciting latent predictions from transformers with the tuned lens}.

\bibitem[{Bostock et~al.(2011)Bostock, Ogievetsky, and Heer}]{2011-d3}
Michael Bostock, Vadim Ogievetsky, and Jeffrey Heer. 2011.
\newblock \href {https://doi.org/10.1109/TVCG.2011.185} {D3: Data-driven documents}.
\newblock \emph{IEEE Trans. Visualization \& Comp. Graphics (Proc. InfoVis)}.

\bibitem[{Brown et~al.(2020)Brown, Mann, Ryder, Subbiah, Kaplan, Dhariwal, Neelakantan, Shyam, Sastry, Askell, Agarwal, Herbert-Voss, Krueger, Henighan, Child, Ramesh, Ziegler, Wu, Winter, Hesse, Chen, Sigler, Litwin, Gray, Chess, Clark, Berner, McCandlish, Radford, Sutskever, and Amodei}]{NEURIPS2020_1457c0d6}
Tom Brown, Benjamin Mann, Nick Ryder, Melanie Subbiah, Jared~D Kaplan, Prafulla Dhariwal, Arvind Neelakantan, Pranav Shyam, Girish Sastry, Amanda Askell, Sandhini Agarwal, Ariel Herbert-Voss, Gretchen Krueger, Tom Henighan, Rewon Child, Aditya Ramesh, Daniel Ziegler, Jeffrey Wu, Clemens Winter, Chris Hesse, Mark Chen, Eric Sigler, Mateusz Litwin, Scott Gray, Benjamin Chess, Jack Clark, Christopher Berner, Sam McCandlish, Alec Radford, Ilya Sutskever, and Dario Amodei. 2020.
\newblock \href {https://proceedings.neurips.cc/paper/2020/file/1457c0d6bfcb4967418bfb8ac142f64a-Paper.pdf} {Language models are few-shot learners}.
\newblock In \emph{Advances in Neural Information Processing Systems}, volume~33, pages 1877--1901. Curran Associates, Inc.

\bibitem[{Clark et~al.(2019)Clark, Khandelwal, Levy, and Manning}]{clark-etal-2019-bert}
Kevin Clark, Urvashi Khandelwal, Omer Levy, and Christopher~D. Manning. 2019.
\newblock \href {https://doi.org/10.18653/v1/W19-4828} {What does {BERT} look at? an analysis of {BERT}{'}s attention}.
\newblock In \emph{Proceedings of the 2019 ACL Workshop BlackboxNLP: Analyzing and Interpreting Neural Networks for NLP}, pages 276--286, Florence, Italy. Association for Computational Linguistics.

\bibitem[{Conmy et~al.(2023)Conmy, Mavor-Parker, Lynch, Heimersheim, and Garriga-Alonso}]{conmy2023automated}
Arthur Conmy, Augustine~N. Mavor-Parker, Aengus Lynch, Stefan Heimersheim, and Adri{\`a} Garriga-Alonso. 2023.
\newblock \href {http://arxiv.org/abs/2304.14997} {Towards automated circuit discovery for mechanistic interpretability}.
\newblock In \emph{Thirty-seventh Conference on Neural Information Processing Systems}.

\bibitem[{Correia et~al.(2019)Correia, Niculae, and Martins}]{correia-etal-2019-adaptively}
Gon{\c{c}}alo~M. Correia, Vlad Niculae, and Andr{\'e} F.~T. Martins. 2019.
\newblock \href {https://doi.org/10.18653/v1/D19-1223} {Adaptively sparse transformers}.
\newblock In \emph{Proceedings of the 2019 Conference on Empirical Methods in Natural Language Processing and the 9th International Joint Conference on Natural Language Processing (EMNLP-IJCNLP)}, pages 2174--2184, Hong Kong, China. Association for Computational Linguistics.

\bibitem[{Dai et~al.(2022)Dai, Dong, Hao, Sui, Chang, and Wei}]{dai-etal-2022-knowledge}
Damai Dai, Li~Dong, Yaru Hao, Zhifang Sui, Baobao Chang, and Furu Wei. 2022.
\newblock \href {https://doi.org/10.18653/v1/2022.acl-long.581} {Knowledge neurons in pretrained transformers}.
\newblock In \emph{Proceedings of the 60th Annual Meeting of the Association for Computational Linguistics (Volume 1: Long Papers)}, pages 8493--8502, Dublin, Ireland. Association for Computational Linguistics.

\bibitem[{Dar et~al.(2023)Dar, Geva, Gupta, and Berant}]{dar-etal-2023-analyzing}
Guy Dar, Mor Geva, Ankit Gupta, and Jonathan Berant. 2023.
\newblock \href {https://doi.org/10.18653/v1/2023.acl-long.893} {Analyzing transformers in embedding space}.
\newblock In \emph{Proceedings of the 61st Annual Meeting of the Association for Computational Linguistics (Volume 1: Long Papers)}, pages 16124--16170, Toronto, Canada. Association for Computational Linguistics.

\bibitem[{Din et~al.(2023)Din, Karidi, Choshen, and Geva}]{din2023jump}
Alexander~Yom Din, Taelin Karidi, Leshem Choshen, and Mor Geva. 2023.
\newblock \href {http://arxiv.org/abs/2303.09435} {Jump to conclusions: Short-cutting transformers with linear transformations}.

\bibitem[{Elhage et~al.(2021)Elhage, Nanda, Olsson, Henighan, Joseph, Mann, Askell, Bai, Chen, Conerly, DasSarma, Drain, Ganguli, Hatfield-Dodds, Hernandez, Jones, Kernion, Lovitt, Ndousse, Amodei, Brown, Clark, Kaplan, McCandlish, and Olah}]{elhage2021mathematical}
Nelson Elhage, Neel Nanda, Catherine Olsson, Tom Henighan, Nicholas Joseph, Ben Mann, Amanda Askell, Yuntao Bai, Anna Chen, Tom Conerly, Nova DasSarma, Dawn Drain, Deep Ganguli, Zac Hatfield-Dodds, Danny Hernandez, Andy Jones, Jackson Kernion, Liane Lovitt, Kamal Ndousse, Dario Amodei, Tom Brown, Jack Clark, Jared Kaplan, Sam McCandlish, and Chris Olah. 2021.
\newblock \href {https://transformer-circuits.pub/2021/framework/index.html} {A mathematical framework for transformer circuits}.
\newblock \emph{Transformer Circuits Thread}.

\bibitem[{Ferrando et~al.(2022)Ferrando, G{\'a}llego, and Costa-juss{\`a}}]{ferrando-etal-2022-measuring}
Javier Ferrando, Gerard~I. G{\'a}llego, and Marta~R. Costa-juss{\`a}. 2022.
\newblock \href {https://doi.org/10.18653/v1/2022.emnlp-main.595} {Measuring the mixing of contextual information in the transformer}.
\newblock In \emph{Proceedings of the 2022 Conference on Empirical Methods in Natural Language Processing}, pages 8698--8714, Abu Dhabi, United Arab Emirates. Association for Computational Linguistics.

\bibitem[{Ferrando and Voita(2024)}]{ferrando_voita2024routes}
Javier Ferrando and Elena Voita. 2024.
\newblock \href {http://arxiv.org/abs/2403.00824} {Information flow routes: Automatically interpreting language models at scale}.

\bibitem[{Geiger et~al.(2021)Geiger, Lu, Icard, and Potts}]{geiger2021causal}
Atticus Geiger, Hanson Lu, Thomas~F Icard, and Christopher Potts. 2021.
\newblock \href {https://openreview.net/forum?id=RmuXDtjDhG} {Causal abstractions of neural networks}.
\newblock In \emph{Advances in Neural Information Processing Systems}.

\bibitem[{Geiger et~al.(2020)Geiger, Richardson, and Potts}]{geiger-etal-2020-neural}
Atticus Geiger, Kyle Richardson, and Christopher Potts. 2020.
\newblock \href {https://doi.org/10.18653/v1/2020.blackboxnlp-1.16} {Neural natural language inference models partially embed theories of lexical entailment and negation}.
\newblock In \emph{Proceedings of the Third BlackboxNLP Workshop on Analyzing and Interpreting Neural Networks for NLP}, pages 163--173, Online. Association for Computational Linguistics.

\bibitem[{Geva et~al.(2022{\natexlab{a}})Geva, Caciularu, Dar, Roit, Sadde, Shlain, Tamir, and Goldberg}]{geva-etal-2022-lm}
Mor Geva, Avi Caciularu, Guy Dar, Paul Roit, Shoval Sadde, Micah Shlain, Bar Tamir, and Yoav Goldberg. 2022{\natexlab{a}}.
\newblock \href {https://doi.org/10.18653/v1/2022.emnlp-demos.2} {{LM}-debugger: An interactive tool for inspection and intervention in transformer-based language models}.
\newblock In \emph{Proceedings of the 2022 Conference on Empirical Methods in Natural Language Processing: System Demonstrations}, pages 12--21, Abu Dhabi, UAE. Association for Computational Linguistics.

\bibitem[{Geva et~al.(2022{\natexlab{b}})Geva, Caciularu, Wang, and Goldberg}]{geva_2022_transformer}
Mor Geva, Avi Caciularu, Kevin~Ro Wang, and Yoav Goldberg. 2022{\natexlab{b}}.
\newblock \href {https://doi.org/10.48550/ARXIV.2203.14680} {Transformer feed-forward layers build predictions by promoting concepts in the vocabulary space}.

\bibitem[{Geva et~al.(2021)Geva, Schuster, Berant, and Levy}]{geva-etal-2021-transformer}
Mor Geva, Roei Schuster, Jonathan Berant, and Omer Levy. 2021.
\newblock \href {https://doi.org/10.18653/v1/2021.emnlp-main.446} {Transformer feed-forward layers are key-value memories}.
\newblock In \emph{Proceedings of the 2021 Conference on Empirical Methods in Natural Language Processing}, pages 5484--5495, Online and Punta Cana, Dominican Republic. Association for Computational Linguistics.

\bibitem[{Hanna et~al.(2023)Hanna, Liu, and Variengien}]{hanna2023does}
Michael Hanna, Ollie Liu, and Alexandre Variengien. 2023.
\newblock \href {http://arxiv.org/abs/2305.00586} {How does gpt-2 compute greater-than?: Interpreting mathematical abilities in a pre-trained language model}.

\bibitem[{Heimersheim and Janiak(2023)}]{docstring}
Stefan Heimersheim and Jett Janiak. 2023.
\newblock \href {https://www.lesswrong.com/posts/u6KXXmKFbXfWzoAXn/a-circuit-for-python-docstrings-in-a-4-layer-attention-only} {The singular value decompositions of transformer weight matrices are highly interpretable}.

\bibitem[{Kaplan et~al.(2020)Kaplan, McCandlish, Henighan, Brown, Chess, Child, Gray, Radford, Wu, and Amodei}]{kaplan2020scaling}
Jared Kaplan, Sam McCandlish, Tom Henighan, Tom~B. Brown, Benjamin Chess, Rewon Child, Scott Gray, Alec Radford, Jeffrey Wu, and Dario Amodei. 2020.
\newblock \href {http://arxiv.org/abs/2001.08361} {Scaling laws for neural language models}.

\bibitem[{Katz and Belinkov(2023)}]{katz-belinkov-2023-visit}
Shahar Katz and Yonatan Belinkov. 2023.
\newblock \href {https://doi.org/10.18653/v1/2023.findings-emnlp.939} {{VISIT}: Visualizing and interpreting the semantic information flow of transformers}.
\newblock In \emph{Findings of the Association for Computational Linguistics: EMNLP 2023}, pages 14094--14113, Singapore. Association for Computational Linguistics.

\bibitem[{Kobayashi et~al.(2020)Kobayashi, Kuribayashi, Yokoi, and Inui}]{kobayashi-etal-2020-attention}
Goro Kobayashi, Tatsuki Kuribayashi, Sho Yokoi, and Kentaro Inui. 2020.
\newblock \href {https://doi.org/10.18653/v1/2020.emnlp-main.574} {Attention is not only a weight: Analyzing transformers with vector norms}.
\newblock In \emph{Proceedings of the 2020 Conference on Empirical Methods in Natural Language Processing (EMNLP)}, pages 7057--7075, Online. Association for Computational Linguistics.

\bibitem[{Kokhlikyan et~al.(2020)Kokhlikyan, Miglani, Martin, Wang, Alsallakh, Reynolds, Melnikov, Kliushkina, Araya, Yan, and Reblitz-Richardson}]{kokhlikyan2020captum}
Narine Kokhlikyan, Vivek Miglani, Miguel Martin, Edward Wang, Bilal Alsallakh, Jonathan Reynolds, Alexander Melnikov, Natalia Kliushkina, Carlos Araya, Siqi Yan, and Orion Reblitz-Richardson. 2020.
\newblock \href {http://arxiv.org/abs/2009.07896} {Captum: A unified and generic model interpretability library for pytorch}.

\bibitem[{Merullo et~al.(2023)Merullo, Eickhoff, and Pavlick}]{merullo2023mechanism}
Jack Merullo, Carsten Eickhoff, and Ellie Pavlick. 2023.
\newblock \href {http://arxiv.org/abs/2305.16130} {A mechanism for solving relational tasks in transformer language models}.

\bibitem[{Miglani et~al.(2023)Miglani, Yang, Markosyan, Garcia-Olano, and Kokhlikyan}]{miglani-etal-2023-using}
Vivek Miglani, Aobo Yang, Aram Markosyan, Diego Garcia-Olano, and Narine Kokhlikyan. 2023.
\newblock \href {https://doi.org/10.18653/v1/2023.nlposs-1.19} {Using captum to explain generative language models}.
\newblock In \emph{Proceedings of the 3rd Workshop for Natural Language Processing Open Source Software (NLP-OSS 2023)}, pages 165--173, Singapore, Singapore. Empirical Methods in Natural Language Processing.

\bibitem[{Nanda and Bloom(2022)}]{nanda2022transformerlens}
Neel Nanda and Joseph Bloom. 2022.
\newblock Transformerlens.
\newblock \url{https://github.com/neelnanda-io/TransformerLens}.

\bibitem[{nostalgebraist(2020)}]{logit_lens}
nostalgebraist. 2020.
\newblock \href {https://www.lesswrong.com/posts/AcKRB8wDpdaN6v6ru/interpreting-gpt-the-logit-lens} {Interpreting gpt: The logit lens}.

\bibitem[{Olsson et~al.(2022)Olsson, Elhage, Nanda, Joseph, DasSarma, Henighan, Mann, Askell, Bai, Chen, Conerly, Drain, Ganguli, Hatfield-Dodds, Hernandez, Johnston, Jones, Kernion, Lovitt, Ndousse, Amodei, Brown, Clark, Kaplan, McCandlish, and Olah}]{olsson2022context}
Catherine Olsson, Nelson Elhage, Neel Nanda, Nicholas Joseph, Nova DasSarma, Tom Henighan, Ben Mann, Amanda Askell, Yuntao Bai, Anna Chen, Tom Conerly, Dawn Drain, Deep Ganguli, Zac Hatfield-Dodds, Danny Hernandez, Scott Johnston, Andy Jones, Jackson Kernion, Liane Lovitt, Kamal Ndousse, Dario Amodei, Tom Brown, Jack Clark, Jared Kaplan, Sam McCandlish, and Chris Olah. 2022.
\newblock \href {https://transformer-circuits.pub/2022/in-context-learning-and-induction-heads/index.html} {In-context learning and induction heads}.
\newblock \emph{Transformer Circuits Thread}.

\bibitem[{OpenAI(2023)}]{openai2023gpt4}
OpenAI. 2023.
\newblock \href {http://arxiv.org/abs/2303.08774} {Gpt-4 technical report}.

\bibitem[{Ouyang et~al.(2022)Ouyang, Wu, Jiang, Almeida, Wainwright, Mishkin, Zhang, Agarwal, Slama, Ray, Schulman, Hilton, Kelton, Miller, Simens, Askell, Welinder, Christiano, Leike, and Lowe}]{ouyang2022training}
Long Ouyang, Jeff Wu, Xu~Jiang, Diogo Almeida, Carroll~L. Wainwright, Pamela Mishkin, Chong Zhang, Sandhini Agarwal, Katarina Slama, Alex Ray, John Schulman, Jacob Hilton, Fraser Kelton, Luke Miller, Maddie Simens, Amanda Askell, Peter Welinder, Paul Christiano, Jan Leike, and Ryan Lowe. 2022.
\newblock \href {http://arxiv.org/abs/2203.02155} {Training language models to follow instructions with human feedback}.

\bibitem[{Radford et~al.(2019)Radford, Wu, Child, Luan, Amodei, and Sutskever}]{radford_language_2019}
Alec Radford, Jeff Wu, Rewon Child, D.~Luan, Dario Amodei, and Ilya Sutskever. 2019.
\newblock \href {https://www.semanticscholar.org/paper/Language-Models-are-Unsupervised-Multitask-Learners-Radford-Wu/9405cc0d6169988371b2755e573cc28650d14dfe} {Language {Models} are {Unsupervised} {Multitask} {Learners}}.

\bibitem[{Sarti et~al.(2023)Sarti, Feldhus, Sickert, and van~der Wal}]{sarti-etal-2023-inseq}
Gabriele Sarti, Nils Feldhus, Ludwig Sickert, and Oskar van~der Wal. 2023.
\newblock \href {https://doi.org/10.18653/v1/2023.acl-demo.40} {Inseq: An interpretability toolkit for sequence generation models}.
\newblock In \emph{Proceedings of the 61st Annual Meeting of the Association for Computational Linguistics (Volume 3: System Demonstrations)}, pages 421--435, Toronto, Canada. Association for Computational Linguistics.

\bibitem[{Stolfo et~al.(2023)Stolfo, Belinkov, and Sachan}]{stolfo2023understanding}
Alessandro Stolfo, Yonatan Belinkov, and Mrinmaya Sachan. 2023.
\newblock \href {http://arxiv.org/abs/2305.15054} {Understanding arithmetic reasoning in language models using causal mediation analysis}.

\bibitem[{Teixeira et~al.()Teixeira, Kelly, Treuille, and Team}]{streamlit}
Thiago Teixeira, Amanda Kelly, Adrien Treuille, and Streamlit Team.
\newblock Streamlit: A faster way to build and share data apps.
\newblock \url{https://streamlit.io/}.

\bibitem[{Tenney et~al.(2020)Tenney, Wexler, Bastings, Bolukbasi, Coenen, Gehrmann, Jiang, Pushkarna, Radebaugh, Reif, and Yuan}]{tenney-etal-2020-language}
Ian Tenney, James Wexler, Jasmijn Bastings, Tolga Bolukbasi, Andy Coenen, Sebastian Gehrmann, Ellen Jiang, Mahima Pushkarna, Carey Radebaugh, Emily Reif, and Ann Yuan. 2020.
\newblock \href {https://doi.org/10.18653/v1/2020.emnlp-demos.15} {The language interpretability tool: Extensible, interactive visualizations and analysis for {NLP} models}.
\newblock In \emph{Proceedings of the 2020 Conference on Empirical Methods in Natural Language Processing: System Demonstrations}, pages 107--118, Online. Association for Computational Linguistics.

\bibitem[{Touvron et~al.(2023{\natexlab{a}})Touvron, Lavril, Izacard, Martinet, Lachaux, Lacroix, Rozière, Goyal, Hambro, Azhar, Rodriguez, Joulin, Grave, and Lample}]{touvron2023llama}
Hugo Touvron, Thibaut Lavril, Gautier Izacard, Xavier Martinet, Marie-Anne Lachaux, Timothée Lacroix, Baptiste Rozière, Naman Goyal, Eric Hambro, Faisal Azhar, Aurelien Rodriguez, Armand Joulin, Edouard Grave, and Guillaume Lample. 2023{\natexlab{a}}.
\newblock \href {http://arxiv.org/abs/2302.13971} {Llama: Open and efficient foundation language models}.

\bibitem[{Touvron et~al.(2023{\natexlab{b}})Touvron, Martin, Stone, Albert, Almahairi, Babaei, Bashlykov, Batra, Bhargava, Bhosale, Bikel, Blecher, Ferrer, Chen, Cucurull, Esiobu, Fernandes, Fu, Fu, Fuller, Gao, Goswami, Goyal, Hartshorn, Hosseini, Hou, Inan, Kardas, Kerkez, Khabsa, Kloumann, Korenev, Koura, Lachaux, Lavril, Lee, Liskovich, Lu, Mao, Martinet, Mihaylov, Mishra, Molybog, Nie, Poulton, Reizenstein, Rungta, Saladi, Schelten, Silva, Smith, Subramanian, Tan, Tang, Taylor, Williams, Kuan, Xu, Yan, Zarov, Zhang, Fan, Kambadur, Narang, Rodriguez, Stojnic, Edunov, and Scialom}]{touvron2023llama2}
Hugo Touvron, Louis Martin, Kevin Stone, Peter Albert, Amjad Almahairi, Yasmine Babaei, Nikolay Bashlykov, Soumya Batra, Prajjwal Bhargava, Shruti Bhosale, Dan Bikel, Lukas Blecher, Cristian~Canton Ferrer, Moya Chen, Guillem Cucurull, David Esiobu, Jude Fernandes, Jeremy Fu, Wenyin Fu, Brian Fuller, Cynthia Gao, Vedanuj Goswami, Naman Goyal, Anthony Hartshorn, Saghar Hosseini, Rui Hou, Hakan Inan, Marcin Kardas, Viktor Kerkez, Madian Khabsa, Isabel Kloumann, Artem Korenev, Punit~Singh Koura, Marie-Anne Lachaux, Thibaut Lavril, Jenya Lee, Diana Liskovich, Yinghai Lu, Yuning Mao, Xavier Martinet, Todor Mihaylov, Pushkar Mishra, Igor Molybog, Yixin Nie, Andrew Poulton, Jeremy Reizenstein, Rashi Rungta, Kalyan Saladi, Alan Schelten, Ruan Silva, Eric~Michael Smith, Ranjan Subramanian, Xiaoqing~Ellen Tan, Binh Tang, Ross Taylor, Adina Williams, Jian~Xiang Kuan, Puxin Xu, Zheng Yan, Iliyan Zarov, Yuchen Zhang, Angela Fan, Melanie Kambadur, Sharan Narang, Aurelien Rodriguez, Robert Stojnic, Sergey Edunov, and Thomas
  Scialom. 2023{\natexlab{b}}.
\newblock \href {http://arxiv.org/abs/2307.09288} {Llama 2: Open foundation and fine-tuned chat models}.

\bibitem[{Vig et~al.(2020)Vig, Gehrmann, Belinkov, Qian, Nevo, Singer, and Shieber}]{causal_mediation_bias}
Jesse Vig, Sebastian Gehrmann, Yonatan Belinkov, Sharon Qian, Daniel Nevo, Yaron Singer, and Stuart Shieber. 2020.
\newblock \href {https://proceedings.neurips.cc/paper_files/paper/2020/file/92650b2e92217715fe312e6fa7b90d82-Paper.pdf} {Investigating gender bias in language models using causal mediation analysis}.
\newblock In \emph{Advances in Neural Information Processing Systems}, volume~33, pages 12388--12401. Curran Associates, Inc.

\bibitem[{Voita et~al.(2023)Voita, Ferrando, and Nalmpantis}]{voita2023neurons}
Elena Voita, Javier Ferrando, and Christoforos Nalmpantis. 2023.
\newblock \href {http://arxiv.org/abs/2309.04827} {Neurons in large language models: Dead, n-gram, positional}.

\bibitem[{Voita et~al.(2019)Voita, Talbot, Moiseev, Sennrich, and Titov}]{voita-etal-2019-analyzing}
Elena Voita, David Talbot, Fedor Moiseev, Rico Sennrich, and Ivan Titov. 2019.
\newblock \href {https://doi.org/10.18653/v1/P19-1580} {Analyzing multi-head self-attention: Specialized heads do the heavy lifting, the rest can be pruned}.
\newblock In \emph{Proceedings of the 57th Annual Meeting of the Association for Computational Linguistics}, pages 5797--5808, Florence, Italy. Association for Computational Linguistics.

\bibitem[{Wang et~al.(2023)Wang, Variengien, Conmy, Shlegeris, and Steinhardt}]{wang2023interpretability}
Kevin~Ro Wang, Alexandre Variengien, Arthur Conmy, Buck Shlegeris, and Jacob Steinhardt. 2023.
\newblock \href {https://openreview.net/forum?id=NpsVSN6o4ul} {Interpretability in the wild: a circuit for indirect object identification in {GPT}-2 small}.
\newblock In \emph{The Eleventh International Conference on Learning Representations}.

\bibitem[{Wei et~al.(2022)Wei, Tay, Bommasani, Raffel, Zoph, Borgeaud, Yogatama, Bosma, Zhou, Metzler, Chi, Hashimoto, Vinyals, Liang, Dean, and Fedus}]{wei2022emergent}
Jason Wei, Yi~Tay, Rishi Bommasani, Colin Raffel, Barret Zoph, Sebastian Borgeaud, Dani Yogatama, Maarten Bosma, Denny Zhou, Donald Metzler, Ed~H. Chi, Tatsunori Hashimoto, Oriol Vinyals, Percy Liang, Jeff Dean, and William Fedus. 2022.
\newblock \href {https://openreview.net/forum?id=yzkSU5zdwD} {Emergent abilities of large language models}.
\newblock \emph{Transactions on Machine Learning Research}.
\newblock Survey Certification.

\bibitem[{Wolf et~al.(2020)Wolf, Debut, Sanh, Chaumond, Delangue, Moi, Cistac, Rault, Louf, Funtowicz, Davison, Shleifer, von Platen, Ma, Jernite, Plu, Xu, Scao, Gugger, Drame, Lhoest, and Rush}]{wolf-etal-2020-transformers}
Thomas Wolf, Lysandre Debut, Victor Sanh, Julien Chaumond, Clement Delangue, Anthony Moi, Pierric Cistac, Tim Rault, Rémi Louf, Morgan Funtowicz, Joe Davison, Sam Shleifer, Patrick von Platen, Clara Ma, Yacine Jernite, Julien Plu, Canwen Xu, Teven~Le Scao, Sylvain Gugger, Mariama Drame, Quentin Lhoest, and Alexander~M. Rush. 2020.
\newblock \href {https://www.aclweb.org/anthology/2020.emnlp-demos.6} {Transformers: State-of-the-art natural language processing}.
\newblock In \emph{Proceedings of the 2020 Conference on Empirical Methods in Natural Language Processing: System Demonstrations}, pages 38--45, Online. Association for Computational Linguistics.

\bibitem[{Zhang et~al.(2022{\natexlab{a}})Zhang, Roller, Goyal, Artetxe, Chen, Chen, Dewan, Diab, Li, Lin, Mihaylov, Ott, Shleifer, Shuster, Simig, Koura, Sridhar, Wang, and Zettlemoyer}]{opt_lm}
Susan Zhang, Stephen Roller, Naman Goyal, Mikel Artetxe, Moya Chen, Shuohui Chen, Christopher Dewan, Mona Diab, Xian Li, Xi~Victoria Lin, Todor Mihaylov, Myle Ott, Sam Shleifer, Kurt Shuster, Daniel Simig, Punit~Singh Koura, Anjali Sridhar, Tianlu Wang, and Luke Zettlemoyer. 2022{\natexlab{a}}.
\newblock \href {https://doi.org/10.48550/ARXIV.2205.01068} {Opt: Open pre-trained transformer language models}.

\bibitem[{Zhang et~al.(2022{\natexlab{b}})Zhang, Roller, Goyal, Artetxe, Chen, Chen, Dewan, Diab, Li, Lin, Mihaylov, Ott, Shleifer, Shuster, Simig, Koura, Sridhar, Wang, and Zettlemoyer}]{zhang2022opt}
Susan Zhang, Stephen Roller, Naman Goyal, Mikel Artetxe, Moya Chen, Shuohui Chen, Christopher Dewan, Mona Diab, Xian Li, Xi~Victoria Lin, Todor Mihaylov, Myle Ott, Sam Shleifer, Kurt Shuster, Daniel Simig, Punit~Singh Koura, Anjali Sridhar, Tianlu Wang, and Luke Zettlemoyer. 2022{\natexlab{b}}.
\newblock \href {http://arxiv.org/abs/2205.01068} {Opt: Open pre-trained transformer language models}.

\end{thebibliography}
\bibliographystyle{acl_natbib}

\appendix

\end{document}